%% file: main.tex
\documentclass[sigconf]{acmart}

\setcopyright{acmlicensed}
\copyrightyear{2025}
\acmYear{2025}
\acmDOI{XXXXXXX.XXXXXXX}
\acmConference[The Web Conference 2025]{The Web Conference 2025}{May 11--16, 2025}{Sydney, Australia}
\acmISBN{XXX-X-XXXX-XXXX-X/XX/XX}  % 更新 ISBN
\acmPrice{}

\usepackage{graphicx} % Required for inserting images
\usepackage{array}

\title{Auto-Drafting Police Reports from Noisy ASR Outputs: A Trust-Centered LLM Approach}

\author{Param Kulkarni}
\authornote{Corresponding author}
\affiliation{
    \institution{Axon}
    \city{Scottsdale}
    \country{AZ USA}
}
\email{pkulkarni@axon.com}

\author{Yingchi Liu}
\affiliation{
    \institution{Axon}
    \city{Scottsdale}
    \country{AZ USA}
}
\email{yliu@axon.com}

\author{Hao-Ming Fu}
\affiliation{
    \institution{Axon}
    \city{Scottsdale}
    \country{AZ USA}
}
\email{hfu@axon.com}

\author{Shaohua Yang}
\affiliation{
    \institution{Axon}
    \city{Scottsdale}
    \country{AZ USA}
}
\email{syang@axon.com}

\author{Isuru Gunasekara}
\affiliation{
    \institution{Axon}
    \city{Scottsdale}
    \country{AZ USA}
}
\email{igunasekara@axon.com}

\author{Matt Peloquin}
\affiliation{
    \institution{Axon}
    \city{Scottsdale}
    \country{AZ USA}
}
\email{mpeloquin@axon.com}

\author{Noah Spitzer-Williams}
\affiliation{
    \institution{Axon}
    \city{Scottsdale}
    \country{AZ USA}
}
\email{noahsw@axon.com}

\author{Xiaotian Zhou}
\affiliation{
    \institution{Worcester Polytechnic Institute}
    \city{Worcester}
    \country{MA USA}
}
\email{xzhou8@wpi.edu}

\author{Xiaozhong Liu}
\authornote{Corresponding author}
\affiliation{
    \institution{Worcester Polytechnic Institute}
    \city{Worcester}
    \country{MA USA}
}
\email{xliu14@wpi.edu}

\author{Zhengping Ji}
\affiliation{
    \institution{Axon}
    \city{Scottsdale}
    \country{AZ USA}
}
\email{zji@axon.com}

\author{Yasser Ibrahim}
\affiliation{
    \institution{Axon}
    \city{Scottsdale}
    \country{AZ USA}
}
\email{yibrahim@axon.com}

\date{December 2024}

\begin{document}

\input{abstract}
\maketitle

\section{Introduction}
\noindent Over the past decade, U.S. law enforcement has faced substantial challenges, including instances where officers have overstepped their authority~\citep{fridell2021relationship, figueroa2012building}. Simultaneously, recent protests from police officers have shed light on their growing vulnerabilities—both from physical threats during law enforcement interactions and from social scrutiny by various organizations~\citep{asquith2021vulnerability, soltes2021occupational}. Additionally, officers spend up to 40\% of their time writing police reports, further exacerbating their workload and detracting from time spent on active policing~\citep{bennett1992national}. These pressures underscore the urgent need to protect the rights of both suspects and officers, and to standardize police case reporting to ensure accountability and due process. Advancements in AI, particularly large language models (LLMs), present a transformative opportunity to address these challenges by  standardizing the documentation of key elements in police reports and making the report writing process more efficient, enhancing both transparency and fairness in the pursuit of justice~\citep{wu2023precedent, gumusel2022annotation}.

\begin{comment}
Over the past decade, U.S. law enforcement has faced substantial challenges, including instances where officers have overstepped their authority or abused suspects without reasonable suspicion or probable cause. Simultaneously, recent protests from police officers have shed light on their growing vulnerabilities—both from physical threats during law enforcement interactions and from social scrutiny by various organizations. These pressures underscore the urgent need to protect the rights of both suspects and officers, and to standardize police case reporting to ensure accountability and due process. Advancements in AI, particularly large language models (LLMs), present a transformative opportunity to address these challenges by automating and standardizing the documentation of key elements in police reports, enhancing both transparency and fairness in the pursuit of justice.
\end{comment}

However, building an LLM-enabled system for police report draft generation presents significant challenges, particularly when dealing with highly noisy automatic speech recognition (ASR) outputs \cite{errattahi2018automatic, noyes1994errors}. These transcripts often contain various forms of distortion and errors, making it difficult to visually identify the critical elements needed for an accurate and reliable draft report. For example, background noise from vehicles, from other people speaking, chatter on the communication devices used by officers. In addition, multiple speakers speaking, especially simultaneously, can cause the transcription to contain potential errors where part of the speech from one speaker is assigned to another speaker. Traditional LLMs can struggle to filter out such interference, raising concerns about the trustworthiness and usefulness of the generated reports  ~\citep{wu2024knowledge}. To overcome these hurdles, careful design approaches are needed to ensure clear instructions about draft quality are defined for the LLM and to measure the accuracy and integrity in the report drafts.  

In this study, we introduce a novel system designed to automatically generate high-quality and trustworthy police report drafts by leveraging ASR devices like body worn cameras (BWC) worn on police officers' bodies. This system integrates an advanced LLM to produce detailed and reliable report drafts. Crucially, to safeguard transparency and to reduce automation bias, our system introduces various safeguards to make sure that a police officer will review and edit the draft before submitting it.

\begin{figure*}
    \centering
    \includegraphics[width=0.8\textwidth]{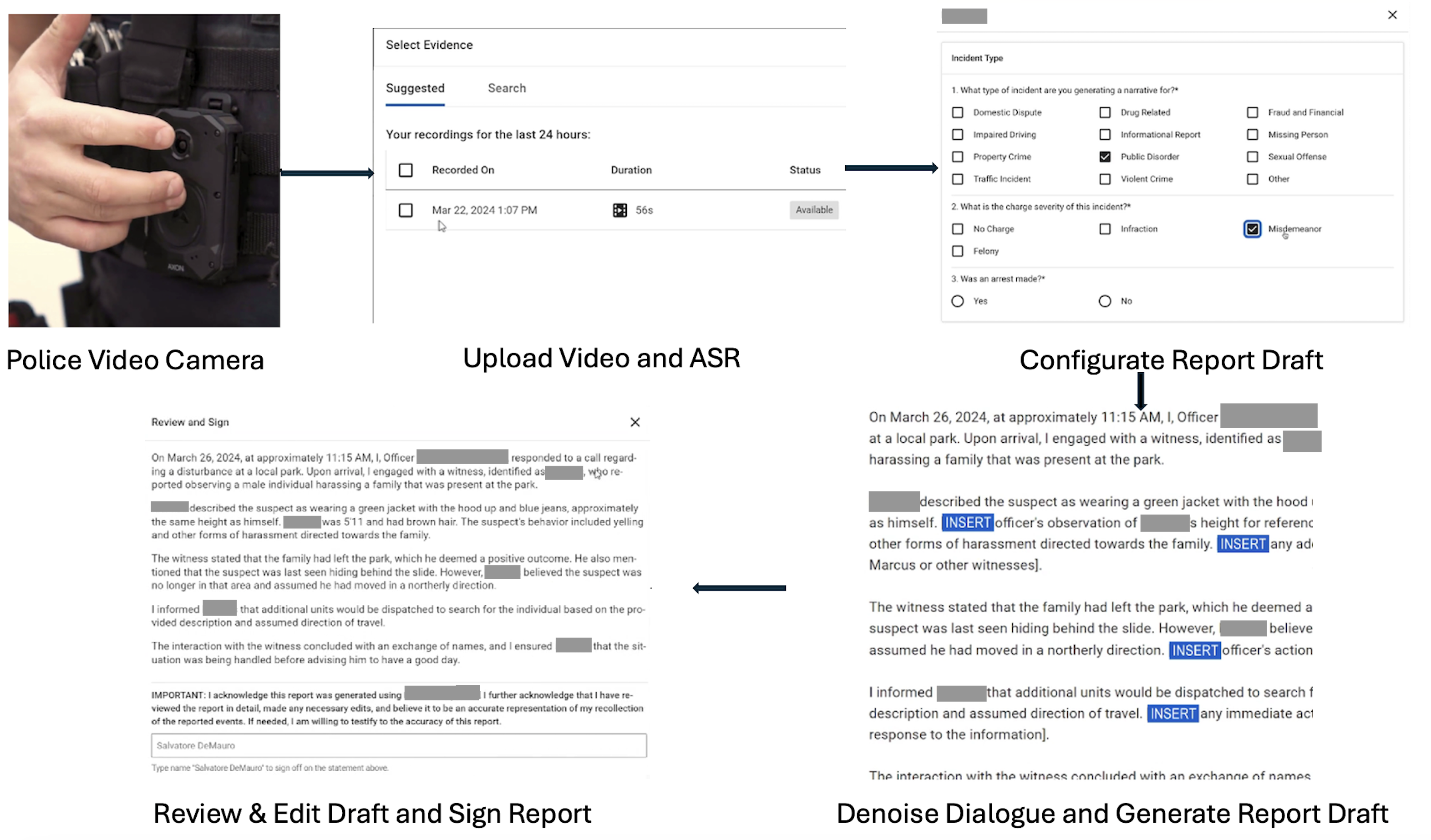} 
    \caption{System Screenshots: Report Draft Generation for Police Officers}
    \label{fig:screen-image}
\end{figure*}

The proposed system’s core functionality includes:

\begin{itemize}
\item \textbf{State-of-the-art LLM with a prompt designed to include safety features:} Capable of synthesizing comprehensive report drafts by efficiently processing noisy, multi-speaker dialogue. 
    \item \textbf{ASR system and LLM Integration:} Enables seamless interaction between ASR devices and the LLM for draft generation during field operations. 
    \item \textbf{Human-in-the-Loop Collaboration:} Facilitates iterative refinement of drafts through collaboration with officers to ensure accuracy, completeness, and contextual alignment.
 
\end{itemize}

This system represents a significant leap toward more transparent, fair, and efficient law enforcement practices by standardizing police reports and enhancing accountability without compromising individual privacy. 

\section{Police Report Draft Generation System}

In this paper, we present a system capable of generating high quality report drafts from noisy dialogue data extracted from police BWC ASR devices. The system captures essential details and events from dialogues involving multiple agents, such as several police officers, witnesses, people of interest, victims, and suspects. Due to the complexity of these interactions and limitations of the BWC device, the data may contain various errors, such as speaker attribution issues and ASR inaccuracies.

\begin{figure*}
    \centering
    \includegraphics[width=0.8\textwidth]{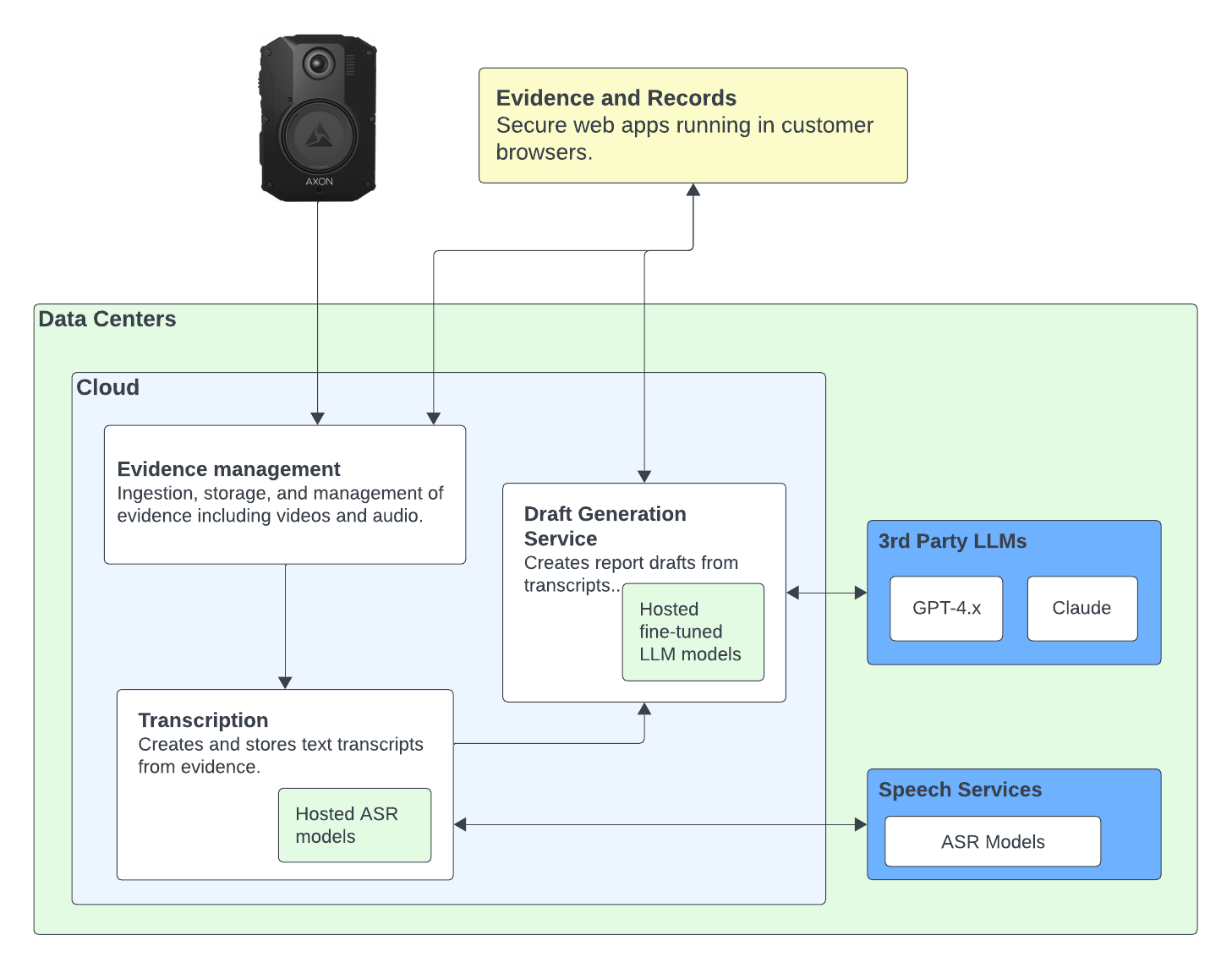} 
    \caption{Draft generation system overview}
    \label{fig:screen-image}
\end{figure*}

Our system addresses these challenges by intelligently processing noisy dialogues and producing coherent, accurate report drafts. Additionally, it incorporates a human-in-the-loop design, allowing for configurable report draft generation parameters. As illustrated in Figure 1, once the BWC footage is uploaded to the system, the officer can start the process of generating the draft. As the first step in human-in-the-loop design, the officer is asked to fill out several details about the incident. To ensure ethical use of LLMs in public safety, the system does not predict the incident type, the severity of charge, and outcomes such as if an arrest was made. The police officers can specify key details such as the type of incident (e.g., "\textit{public disorder}" or "\textit{sexual offense}") and the severity of charges (e.g., "\textit{misdemeanor}" or "\textit{felony}"). These details do not play any role in generating the draft using the LLM. This done by design to ensure that the draft is only an accurate reflection of the evidence in the transcript and does not make conclusory statements about the suspect.

Followed by entering details about the incident, the officer can generate the draft as shown in Figure 1. As a second step in human-in-the-loop design, the draft is designed to contain several INSERT statements. These statements are used by the model when more information is needed beyond what is found in the transcript. This feature allows officers to easily insert important details, such as specific actions and observations that may not have been captured accurately by the ASR device. The system requires the officer to edit each of these statements before proceeding to the next step. The system will not let the officer proceed to the next step until all the INSERT statements are edited. This offers ample opportunity and incentive for officers to read the drafts and making edits beyond what is suggested by the model. As the final step in human-in-the-loop design, the officer is required to sign the edited draft with their full name before they can submit it. These several steps in our system design which require officer input before proceeding are designed to minimize automation bias and quality of police reports.

Our system, as shown in Figure 2, uses a highly secure digital evidence management service to store the video footage from the BWC device. The system is capable of interfacing with several generations of BWC devices. The audio from this evidence is then passed to our ASR service. The system is designed to work with several hosted ASR models. The output of the ASR engine is then passed to an LLM service capable of generating the draft which are then securely transferred to the customer's record management system. Draft generation is powered by a carefully designed prompt which includes detailed instructions about draft quality as well as safety requirements. Once the draft is generated, it is delivered to the officer for review, which then results in them editing and submitting the draft after going through the safeguards mentioned above. The system is designed to be flexible enough to interact with various types of language models and ASR systems. This provides flexibility in updating the system with models with the best performance on quality and safety metrics. This system is designed to significantly improves the efficiency of police report generation, while ensuring the final report is both trustworthy and comprehensive.
% \begin{figure*}[h]
%     \centering
%     \includegraphics[width=\textwidth]{toy_example_version_7.pdf} 
%     \caption{Automated Draft Generation from Body Camera ASR(Noisy Dialog) : A Three-Stage System for Denoising, Event Detection, and Report Summarization}
%     \label{fig:pdf-image}
% \end{figure*}

% \begin{figure*}[h]
%     \centering
%     \includegraphics[width=0.7\textwidth]{system_architecture.png} 
%     \caption{System Architecture}
%     \label{fig:system-image}
% \end{figure*}

% We need system screenshots(Figure~\ref{fig:screen-image}):

% For dialogue generation, we used a specific prompt: 
% \begin{lstlisting}[language=, frame=single, breaklines=true, backgroundcolor=\color{gray!10}, basicstyle=\ttfamily]
% I will provide you with a JSON data of an event extraction from caselaw. Please imagine the scene at the time of the incident and simulate a conversation between the parties involved. Return only the conversation to me without any narration. The length should be around {} sentences, in English. Simulate only the scene at the time of the incident, not a courtroom dialogue.\end{lstlisting} 

\section{Usability Evaluation with Police Officers}
To assess the practical effectiveness, reliability, and user-friendliness of our police report draft generation system, we conducted a usability evaluation involving active police officers. The primary objective of this evaluation was to ensure that the system meets the specific needs of law enforcement professionals, integrates seamlessly into their workflows, and produces accurate, actionable report drafts.

\begin{table*}
\centering
\begin{tabular}{|c|c|c|c|c|c|c|}
\hline
\textbf{Average scores for each category} & \textbf{Overall} & \textbf{Completeness} & \textbf{Neutrality} & \textbf{Objectivity} & \textbf{Terminology} & \textbf{Coherence} \\ \hline
\textbf{Reports written with no system use} & 3.93 & 3.73 & 4.12 & 4.11 & 3.97 & 3.75 \\ \hline
\textbf{Reports written system assistance} & 4.06 & 3.83 & 4.13 & 4.12 & 4.20 (p=0.033) & 4.05 (p=0.019) \\ \hline
\end{tabular}
\caption{System effectiveness study}
\end{table*}

We found that the preliminary system already has a strong positive impact on officer's workflow, as indicated by several usability metrics and surveys of first-line police officers. On average, users have assigned a rating of 4.46 on a scale of 1 to 5 for drafts generated by this system, and claimed that the system helped save 21.93 minutes on writing per report, or 41.81\% of a usual report writing time. These demonstrate that the system has been widely adopted by first-line officers, has created positive impact on the workflow, and received highly positive feedback from police officers.

In addition, for a systematic evaluation, we conducted an expert led double-blind study on 113 pairs of reports written with and without the assistance of the system. The 24 experts for this study were chosen based on their background in law enforcement equity and inclusion. The evaluation covers a wide range of aspects, including report completeness, neutrality, objectivity, terminology, and coherence. Results show that reports written with assistance of the system statistically significantly outperform those written without any assistance on aspects terminology (4.20 vs. 3.97) and coherence (4.05 vs. 3.75), while performing comparably on completeness, neutrality, and objectivity, as elaborated in Table 1.

\section{Conclusion}
In this work, we propose a human-in-the-loop system for generating trustworthy police case report drafts, which holds significant potential for enhancing the integrity and transparency of law enforcement processes. Currently being piloted with 326 police agencies, the initial feedback has been promising. Police officers are benefiting from this AI-driven innovation, as it streamlines and standardizes the report generation process, while also safeguarding the rights of suspects. This system represents a vital step forward in ensuring accountability and trust in law enforcement through more efficient and reliable documentation.

\newpage
\bibliographystyle{ACM-Reference-Format}
\bibliography{reference}
\end{document}

%% file: abstract.tex
\begin{abstract}
Achieving a delicate balance between fostering trust in law enforcement and protecting the rights of both officers and civilians continues to emerge as a pressing research and product challenge in the world today. In the pursuit of fairness and transparency, this study presents an innovative AI-driven system designed to generate police report drafts from complex, noisy, and multi-role dialogue data. Our approach intelligently extracts key elements of law enforcement interactions and includes them in the draft, producing structured narratives that are not only high in quality but also reinforce accountability and procedural clarity. This framework holds the potential to transform the reporting process, ensuring greater oversight, consistency, and fairness in future policing practices. A demonstration video of our system can be accessed at \textbf{https://drive.google.com/file/d/1kBrsGGR8e3B5xPSblrchRGj-Y-kpCHNO/view?usp=sharing}
\end{abstract}